\documentclass{edm_article}
\usepackage{url}
\begin{document}

\title{Exploring the Capabilities of Prompted Large Language Models in Educational and Assessment Applications}

\numberofauthors{3}
\author{
\alignauthor
Subhankar Maity\\
       \affaddr{IIT Kharagpur, India}\\
       \email{subhankar.ai@kgpian.\\iitkgp.ac.in}
\alignauthor
Aniket Deroy\\
    \affaddr{IIT Kharagpur, India}\\
    \email{roydanik18@kgpian.\\iitkgp.ac.in} 
\alignauthor
Sudeshna Sarkar\\
    \affaddr{IIT Kharagpur, India}\\
   \email{sudeshna@cse.iitkgp.ac.in}
 }

\maketitle

\begin{abstract}
In the era of generative artificial intelligence (AI), the fusion of large language models (LLMs) offers unprecedented opportunities for innovation in the field of modern education. We embark on an exploration of prompted LLMs within the context of educational and assessment applications to uncover their potential. Through a series of carefully crafted research questions, we investigate the effectiveness of prompt-based techniques in generating open-ended questions from school-level textbooks, assess their efficiency in generating open-ended questions from undergraduate-level technical textbooks, and explore the feasibility of employing a \textit{chain-of-thought} inspired multi-stage prompting approach for language-agnostic multiple-choice question (MCQ) generation. Additionally, we evaluate the ability of prompted LLMs for language learning, exemplified through a case study in the low-resource Indian language Bengali, to explain Bengali grammatical errors. We also evaluate the potential of prompted LLMs to assess human resource (HR) spoken interview transcripts. By juxtaposing the capabilities of LLMs with those of human experts across various educational tasks and domains, our aim is to shed light on the potential and limitations of LLMs in reshaping educational practices.
\end{abstract}

\keywords{Educational, Large language models, Prompt, Question Generation, Assessment} 

\section{Introduction}
In the current era of rapid technological advancement, the integration of generative AI models, particularly LLMs, represents a pivotal shift in educational practices and assessment methodologies \cite{r1,r3,r4}. These LLMs, driven by generative AI, have a profound grasp of natural language and formidable computational prowess, offering promising transformative potential in both learning facilitation and student evaluation \cite{r2}. Our study embarks on a thorough exploration of the utilization of LLMs across a spectrum of educational and assessment contexts, with a focus on elucidating their efficacy and identifying areas ready for improvement. Our objective is to address key research questions, striving to unveil the multifaceted potential of LLMs while acknowledging the inherent complexities and challenges in their integration. 

Our investigation is underscored by the use of a prompting approach, aimed at enhancing the capabilities of LLMs in subsequent tasks by providing additional information, known as a "prompt", to guide their generation process \cite{r5}. Recently, the use of prompts has gained significant attention across different natural language generation tasks, such as summarization \cite{r6}, machine translation \cite{r7}, etc. Through rigorous examination and analysis, our objective is to contribute meaningfully to the ongoing discourse surrounding the integration of generative AI models in education, providing nuanced insights that inform future research endeavors and educational practices.

\section{Related Work}
The present research studies \cite{r23, r24} explore various prompt-based strategies for question generation (QG). \cite{r23} curate the KHANQ dataset, categorizing each data sample into a triple of <Context, Prompt, Question> and investigate prompt-based QG using LLMs such as BERT generation \cite{r26}, BART \cite{r13}, GPT2 \cite{r27}, T5 \cite{r12}, and UniLM \cite{r28}. The prompts used in KHANQ are tailored according to the learners' background knowledge and comprehension of the subject matter. Despite the considerable value of the KHANQ dataset, the authors have not made it available so far. \cite{r24} utilize prompt-based fine-tuning to formulate multi-hop questions. The methodology entails a sequence of tasks, beginning with QG and subsequently transitioning to question-answering (QA), which is iteratively performed to refine the QG process. T5 is used to train both the QG and QA models. Additionally, question paraphrasing is implemented to enhance the method's robustness. Lastly, prompt-based fine-tuning is employed to produce high-quality questions. They generated a prompt by selecting pertinent words related to the accurate answer and evaluated their model on the HotpotQA \cite{r30}, SQuAD \cite{r29}, and Quora Question Pairs datasets \cite{r31}. 

Recent studies \cite{r32, r33, r24} in automated QG that leverage LLMs have utilized single-hop QA datasets such as SQuAD and multi-hop QA datasets such as HotpotQA. These QA datasets comprise <Context, Question, Answer> triples, wherein \textit{Context} denotes a contextual document, \textit{Question} is a query formulated by a human and \textit{Answer} is its associated response. Current QG methods have also benefited from the availability of QA datasets, such as the Natural Questions corpus \cite{r35}, QuAC \cite{r36}, TriviaQA \cite{r34}, NewsQA \cite{r38}, QG-STEC \cite{r37}, etc. However, it is worth noting that we have identified several limitations in the existing datasets:

\begin{itemize}
    \item[-] As highlighted in \cite{r39, r40}, these datasets are limited in the generation of simple factual questions.

    \item[-] As indicated by \cite{r39, r41, r42}, 
     many of these QA datasets are sourced from open-access platforms, such as Wikipedia articles. Typically, they lack the educational aspect and do not require higher-order cognitive skills to answer them.
     
\end{itemize}

Moreover, there has been no exploration of the capabilities of prompted LLMs for generating open-ended questions from educational textbooks.

The generation of MCQs in multilingual settings, particularly for low-resource languages, is crucial to overcome language barriers, improve accessibility, and advance education in marginalized communities. Although previous research \cite{r44} has been conducted for the English language, involving fine-tuning of a T5 model on the DG-RACE dataset \cite{r45} to produce distractors for MCQ, currently there is no research available for multilingual contexts, such as German, Hindi, and Bengali, where an encoder-decoder-based model is used for distractor generation. Additionally, there is currently no research on MCQ generation that investigates the potential of the \textit{chain-of-thought} \cite{r9} inspired prompt-based method to generate MCQs in various languages.

Despite increasing interest in grammatical error correction (GEC) and the availability of GEC datasets in major languages such as English \cite{r46, r47, r48}, Chinese \cite{r50}, German \cite{r51}, Russian \cite{r52}, Spanish \cite{r53}, etc., there is a noticeable shortage of real-world GEC datasets specifically designed for low-resource languages such as Bengali (despite being the $7^{th}$ most spoken language worldwide \cite{r49}). Current synthetic Bengali GEC datasets, as mentioned in \cite{r54}, lack the authenticity and diversity required to represent the complexities of real-world language usage. Although there is existing GEC research for Bengali \cite{r54, r55, r56, r57, r58}, no effort has been made in the domains of feedback or explanation generation within this particular context. Furthermore, there has been no investigation in GEC to assess the potential of generative pre-trained LLMs like GPT-4 Turbo, GPT-3.5 Turbo, Llama-2, etc., for low-resource languages such as Bengali.

Recent studies have explored aspects of speech scoring, such as assessing response content. This involves modeling features extracted from response transcriptions alongside the corresponding question to gauge response relevance \cite{r60, r61}. Expanding on this, \cite{r62} improved their approach by integrating acoustic cues and grammar features to enhance scoring accuracy. In a more recent investigation, \cite{r63} used speech and text transformers \cite{r64} to evaluate candidate speech. To our knowledge, no research has investigated the use of state-of-the-art LLMs for automated human resource (HR) interview evaluation. Moreover, earlier research in automated speech scoring focused primarily on \textit{scoring}, with minimal emphasis on \textit{error detection} and \textit{providing feedback along with suggestions} for improvement.

\section{Research Questions}

In this section, we present the key research questions guiding our investigation of the capabilities of prompted LLMs across diverse educational and assessment contexts. These research questions serve as focal points, with the aim of evaluating the effectiveness of LLMs compared to human experts in different tasks and domains. We address the following research questions (RQs) on a diverse set of educational topics, as described below.

\begin{itemize}
    \item[\textbf{RQ1:}] To what extent are prompt-based techniques \cite{r23, r24} effective in generating open-ended questions using LLMs from school-level textbooks compared to human experts?

    \item[\textbf{RQ2:}] To what extent are prompt-based techniques effectively enabling LLMs to generate open-ended questions from undergraduate-level technical textbooks compared to human experts?
 
    \item[\textbf{RQ3:}] Can a \textit{chain-of-thought} \cite{r9} inspired multi-stage prompting approach be developed to generate language-agnostic multiple-choice questions using GPT-based models?

    \item[\textbf{RQ4:}] To what extent are pre-trained LLMs capable of explaining Bengali grammatical errors compared to human experts?

    \item[\textbf{RQ5:}] How ready are pre-trained LLMs to assess human resource spoken interview transcripts compared to human experts?
\end{itemize}

\section{Current Research Progress}

In this section, we discuss the current research progress that has been made in addressing the aforementioned research questions.

\textit{\textbf{RQ1: To what extent are prompt-based techniques \cite{r23, r24} effective in generating open-ended questions using LLMs from school-level textbooks compared to human experts?}}

To answer this research question, we propose to examine the efficacy of prompt-based methods \cite{r23, r24} in generating open-ended questions using LLMs from school-level textbooks, compared to human experts. Prompt-based techniques entail furnishing textual cues or prompts to guide LLMs in crafting questions coherent with a given context. These prompts act as signals for the LLMs to produce relevant and coherent questions. Our study aims to investigate the effectiveness of these prompt-based techniques in generating descriptive and reasoning-based questions tailored to educational contexts. 

In our methodology \cite{r70}, we address the challenge posed by the inadequacy of existing QA datasets for prompt-based QG in educational settings by curating a new dataset called EduProbe. This dataset is specifically adapted for school-level subjects (e.g.,  history, geography, economics, environmental studies, and science) and draws on the rich content of the NCERT\footnote{NCERT: \url{http://tinyurl.com/3x7hm2jk}} textbooks. Each instance in the dataset is annotated with quadruples comprising: 1) \textit{Context}: a segment serving as the basis for question formulation, 2) \textit{Long Prompt}: an extended textual cue encompassing the core theme of the context, 3) \textit{Short Prompt}: a condensed representation of crucial information or focus within the context, and 4) \textit{Question}: a question in line with the context and aligned with the prompts. 

Different prompts not only speed up the process of creating questions but also improve the overall quality and diversity of the questions produced by providing LLMs additional guidance on what information to give more importance when creating questions. We explore various prompt-based QG techniques (e.g., \textit{long prompt}, \textit{short prompt}, and \textit{without prompt}) by fine-tuning pre-trained transformer-based LLMs, including PEGASUS \cite{r11}, T5 \cite{r12}, and BART \cite{r13}. Furthermore, we examine the performance of two general-purpose pre-trained LLMs, text-davinci-003 \cite{r14} and GPT-3.5 Turbo, using a zero-shot prompting approach. 

Through automated evaluation, we demonstrate that T5 (\textit{with long prompt}) outperforms other LLMs, although it falls short of the human baseline. Intriguingly, text-davinci-003 consistently shows superior results compared to other LLMs in various prompt settings, even surpassing them in human evaluation criteria. However, prompt-based QG models mostly fall below the human baseline, indicating the need for further exploration and refinement in this domain.

\textit{\textbf{RQ2: To what extent are prompt-based techniques effectively enabling LLMs to generate open-ended questions from undergraduate-level technical textbooks compared to human experts?}}

To address this research question, we delve into the effectiveness of prompt-based techniques in facilitating LLMs to generate open-ended questions from technical textbooks at the undergraduate level, in comparison to human experts. Our investigation focuses on the automated generation of various open-ended questions in the technical domain, an area that is relatively less explored in educational QG research \cite{r39}. 

To facilitate our study, we curate EngineeringQ from undergraduate level technical textbooks on subjects such as operating systems and computer networks. This dataset is designed for prompt-based QG and comprises triples consisting of 1) \textit{Context}: segments from which questions are derived, 2) \textit{Prompt}: concise and specific keyphrases guiding QG, and 3) \textit{Question}: questions coherent with context and prompt. 

We evaluate several fine-tuned encoder-decoder based LLMs, such as Pegasus, BART, Flan-T5 \cite{r17}, and T5, on EngineeringQ. Additionally, we explore the potential of general-purpose decoder-only LLMs like GPT-3.5 Turbo, text-davinci-003, and GPT-4 \cite{r16} using a zero-shot prompting approach. Our assessment involves both automated metrics and human evaluation by domain experts. Moreover, we examine the domain adaptation \cite{r68, r39, r69} capability of LLMs by fine-tuning the best-performing LLM on school-level subjects (e.g.,  history, geography, economics, environmental studies, and science) and assessing its efficacy on undergraduate-level computer science and information technology subjects (e.g., operating systems and computer networks) for zero-shot and few-shot QG. To gauge question complexity, we employ Bloom’s revised taxonomy \cite{r59}, enhancing our understanding of their educational significance. 

Experimental findings indicate that T5\textsubscript{LARGE} outperforms other LLMs in automated evaluation metrics, while text-davinci-003 excels in human evaluation metrics. However, LLMs in both scenarios fall short of the human baseline, highlighting the need for further refinement and exploration in this domain.

\textit{\textbf{RQ3: Can a \textit{chain-of-thought} \cite{r9} inspired multi-stage prompting approach be developed to generate language-agnostic multiple-choice questions using GPT-based models?}}

To answer this research question, we present a novel \textit{chain-of-thought} inspired multi-stage prompting strategy for crafting language-agnostic MCQs utilizing GPT-based models \cite{r71}. This method, known as the multi-stage prompting approach (MSP), capitalizes on the strengths of GPT models such as text-davinci-003 and GPT-4, renowned for their proficiency across diverse natural language processing tasks. Our proposed MSP technique integrates the innovative concept of \textit{chain-of-thought} prompting \cite{r9}, wherein the GPT model receives a sequence of interconnected cues to guide the MCQ generation process. 

We evaluated our proposed language-agnostic MCQ generation method on several datasets across different languages. SQuAD served as the MCQ generation dataset for English (En), while GermanQuAD \cite{r65} was utilized for German (De). For generating questions in Hindi (Hi), we employed HiQuAD \cite{r66}, and for Bengali (Bn), we utilized BanglaRQA \cite{r67}.

Through automated evaluation, we consistently demonstrate the superiority of the MSP method over the conventional single-stage prompting (SSP) baseline, evident in the production of high-quality distractors crucial for effective MCQs. Furthermore, our one-shot MSP method enhances automatic evaluation results, contributing to improved distractor generation in multiple languages, including English, German, Bengali, and Hindi. In human evaluation, questions generated using our proposed MSP approach exhibit superior levels of \textit{grammaticality} \cite{r18}, \textit{answerability} \cite{r18}, and \textit{difficulty} \cite{r19} for high-resource languages (e.g., En, De), underscoring its effectiveness in diverse linguistic contexts. However, further research and fine-tuning of GPT-based models might be required to improve the results for low-resource languages (e.g., Hi, Bn) and to reduce the disparity with high-resource languages (e.g., En, De) in both automated and human evaluation criteria.

\textit{\textbf{RQ4: To what extent are pre-trained LLMs capable of explaining Bengali grammatical errors compared to human experts?}}

GEC tools, driven by advanced generative AI, excel at rectifying linguistic inaccuracies in user input. However, they often lack in furnishing essential natural language explanations, crucial for language learning and comprehension of grammatical rules. Particularly in low-resource Indian languages like Bengali, there is limited exploration of these tools, necessitating grammatical error explanation (GEE) systems that not only correct sentences but also provide explanations for errors. 

To address this research question, we propose an investigation into the proficiency of pre-trained LLMs including GPT-4 Turbo, GPT-3.5 Turbo, text-davinci-003, text-babbage-001, text-curie-001, text-ada-001, llama-2-7b \cite{r15}, llama-2-13b, and llama-2-70b in explaining Bengali grammatical errors compared to human experts. 

We introduce a real-world, multi-domain dataset sourced from various domains such as Bengali essays, social media, and news, serving as an evaluation benchmark for the GEE system. This dataset facilitates the assessment of various pre-trained LLMs against human experts for performance comparison in a one-shot prompt setting. 

Our methodical experimental procedure involved both LLM and human experts, performing two crucial tasks independently. First, they were tasked with producing an accurate Bengali sentence by detecting and correcting errors in the provided sentences, ensuring both grammatical correctness and contextual appropriateness. Second, for each corrected error, they were required to categorize the \textit{error type} and offer concise \textit{explanations} concerning the grammatical, syntactical, or semantic issues addressed.

Our research highlights the limitations in the automatic deployment of current state-of-the-art pre-trained LLMs for Bengali GEE. We advocate for human intervention, proposing the integration of manual checks to refine GEC tools in Bengali, emphasizing the educational aspect of language learning.

\textit{\textbf{RQ5: How ready are pre-trained LLMs to assess human resource spoken interview transcripts compared to human experts?}}

To address this research question, we propose a detailed examination of the readiness of pre-trained LLMs in evaluating human resource (HR) spoken interview transcripts compared to human experts. Our comprehensive analysis encompasses a range of prominent pre-trained LLMs, including GPT-4 Turbo, GPT-3.5 Turbo, text-davinci-003, text-babbage-001, text-curie-001, text-ada-001, llama-2-7b, llama-2-13b, and llama-2-70b, assessing their performance in \textit{providing scores}, \textit{error identification}, and \textit{offering feedback and improvement suggestions} to candidates during simulated HR interviews. 

We introduce a dataset named HURIT (\underline{Hu}man \underline{R}esource \underline{I}nterview \underline{T}ranscripts), comprising HR interview transcripts collected from real-world scenarios. The dataset consists of HR interview transcripts obtained from L2 English speakers, primarily featuring interviews conducted in the Asian region. These transcripts are derived from simulated HR interviews in which students provided their responses. The responses were captured in \textit{.mp3} format and subsequently transcribed into text using OpenAI's Whisper large-v2 model \cite{r20}. This dataset facilitates the evaluation of various pre-trained LLMs against human experts for performance comparison in a zero-shot prompt setting. 

Our approach involved a structured assessment procedure in which both LLMs and human assessors independently \textit{scored}, \textit{identified errors}, and \textit{provided constructive feedback} on HR interview transcripts. This comprehensive method enabled a thorough evaluation of each LLM's performance, encompassing its \textit{scoring accuracy}, \textit{error detection}, and \textit{feedback provision}. Additionally, we compared their abilities with those of expert human evaluators on various human evaluation criteria, such as \textit{fluency}, \textit{coherence}, \textit{tone/politeness}, \textit{relevance}, \textit{conciseness}, and \textit{grammaticality} \cite{r21, r22}.

Our findings highlight the proficiency of pre-trained LLMs, particularly GPT-4 Turbo and GPT-3.5 Turbo, in delivering evaluations comparable to those provided by expert human evaluators. However, while these LLMs excel in \textit{scoring} candidates, they often struggle to \textit{identify errors} and \textit{provide actionable feedback} for performance improvement in HR interviews. Our research underscores that although pre-trained LLMs demonstrate promise in certain aspects, they are not yet fully equipped for automatic deployment in HR interview assessments. Instead, we advocate for a human-in-the-loop approach, emphasizing the importance of manual checks to address inconsistencies and improve the quality of feedback provided, presenting a more viable strategy for HR interview assessment.

\section{Conclusion}
Our study addressed key research questions about the integration of LLMs in educational and assessment applications. We investigated the effectiveness of prompt-based techniques in generating open-ended questions from school-level textbooks using LLMs, highlighting promising but imperfect performance compared to human experts. Despite advancements, LLMs struggled to match human expertise in generating open-ended questions from undergraduate-level technical textbooks, indicating areas for improvement. Additionally, our proposed MSP approach for crafting language-agnostic MCQs shows that further research and fine-tuning of GPT models are required to improve the results in low-resource languages (e.g., Hi, Bn). Furthermore, our exploration of the ability of LLMs to \textit{explain} Bengali grammatical errors revealed deficiencies, underscoring the importance of human intervention. Lastly, while LLMs showed competence in \textit{scoring} HR interview transcripts, they encountered challenges in \textit{error identification} and \textit{feedback provision}, emphasizing the need for human oversight. Overall, our study underscores the potential of LLMs in educational and assessment applications but highlights the ongoing need for research and refinement to fully harness their capabilities. 

In the doctoral consortium, we anticipate receiving recommendations and feedback regarding the current status of our research progress.
\bibliographystyle{abbrv}
\bibliography{sigproc}  
%
\end{document}